\documentclass[fleqn,10pt]{wlscirep}
\usepackage[utf8]{inputenc}
\usepackage[T1]{fontenc}
\usepackage{multirow}
\usepackage{comment}
\usepackage{amsmath}
\usepackage{bm}
\usepackage{graphicx}
\usepackage{subcaption}
\usepackage{hyperref}
\usepackage[normalem]{ulem}

\title{Feature Impact Analysis on Top Long-Jump Performances with Quantile Random Forest and Explainable AI Techniques}

\author[1,*]{Qi Gan}
\author[1]{Stephan Clémençon}
\author[2]{Mounîm A.El-Yacoubi}
\author[3]{Sao Mai Nguyen}
\author[4]{Eric Fenaux}
\author[1]{Ons Jelassi}
\affil[1]{LTCI, Télécom Paris, Institut Polytechnique de Paris, 91120 Palaiseau, France}
\affil[2]{SAMOVAR, Télécom SudParis, Institut Polytechnique de Paris, 91120 Palaiseau, France}
\affil[3]{U2IS, ENSTA Paris, Institut Polytechnique de Paris, 91120 Palaiseau, France}
\affil[4]{Ef-e-science, 75000 Paris, France}

\affil[*]{qi.gan@telecom-paris.fr}

\begin{abstract}
Biomechanical features have become important indicators for evaluating athletes’ techniques. Traditionally, experts propose significant features and evaluate them using physics equations. However, the complexity of the human body and its movements makes it challenging to explicitly analyze the relationships between some features and athletes’ final performance. With advancements in modern machine learning and statistics, data analytics methods have gained increasing importance in sports analytics. In this study, we leverage machine learning models to analyze expert-proposed biomechanical features from the finals of long jump competitions in the World Championships. The objectives of the analysis include identifying the most important features contributing to top-performing jumps and exploring the combined effects of these key features. Using quantile regression, we model the relationship between the biomechanical feature set and the target variable (effective distance), with a particular focus on elite-level jumps. To interpret the model, we apply SHapley Additive exPlanations (SHAP) alongside Partial Dependence Plots (PDPs) and Individual Conditional Expectation (ICE) plots. The findings reveal that, beyond the well-documented velocity-related features, specific technical aspects also play a pivotal role. For male athletes, the angle of the knee of the supporting leg before take-off is identified as a key factor for achieving top 10\% performance in our dataset, with angles greater than 169° contributing significantly to jump performance. In contrast, for female athletes, the landing pose and approach step technique emerge as the most critical features influencing top 10\% performances, alongside velocity. This study establishes a framework for analyzing the impact of various features on athletic performance, with a particular emphasis on top-performing events.

\end{abstract}
\begin{document}

\flushbottom
\maketitle
\thispagestyle{empty}

%\noindent Please note: Abbreviations should be introduced at the first mention in the main text – no abbreviations lists. Suggested structure of main text (not enforced) is provided below.

\section*{Introduction}

Feature-based analysis to evaluate athlete techniques is widely adopted in track-and-field sports analytics. Generally, the final performance (e.g., official distance in long jump) provides an oversimplified characterization of athletes' performances, offering limited guidance on how to enhance their techniques. To overcome this limitation, researchers have utilized additional features (i.e. biomechanical metrics in this work) based on their expertise to better evaluate the detailed performance of the athletes. For instance, in long jump, experts have proposed features that describe the dynamics of the last few steps before take-off, such as velocities, step lengths and step times, as well as features that depict poses during the step-on-board and landing phases, such as body inclination angle and trunk inclination angle \cite{Ref_report09, Ref_report17men, Ref_report17women, Ref_report18men, Ref_report18women}.

The main goal of this work is to study feature-based analysis in sports, using long jump as a case study. In this work, we refer to feature analysis as a two-fold task. The first task is to evaluate the contribution of different features to the final performance by identifying the most important ones. The second task is to uncover the interactions between the features, which helps provide insight into the mechanism and offers effective guidance to real-world practice. To accomplish these goals, extensive research has been undertaken with methods including descriptive biomechanics\cite{campos2013three, mendoza2011biomechanical}, physical modeling\cite{ramey1983biomechanics, seyfarth1999dynamics, shang2022research}, statistical evaluation\cite{panoutsakopoulos2021biomechanical, kozlova2020individual, dos2023post} and machine learning modeling\cite{chmait2021artificial, uccar2022using, odong2023introduction, carvalho2024swimming, lalwani2022machine}. 

Descriptive biomechanics has been used in applied research studies and technical reports on competitions. Campos et al.\cite{campos2013three} studied individual patterns descriptively with the features from the long jump finals at the 2008 World Championships. Mendoza et al.\cite{mendoza2011biomechanical} confirmed the recent technique models of long jump with descriptive studies on the long jump features from the 2009 World Championships. This method, while effective for deriving qualitative conclusions by incorporating expert knowledge, is unable to quantify the influence of features on the target variable. Moreover, these studies largely depend on expert knowledge and are therefore difficult to generalize to different domains and applications. 

To facilitate quantitative studies, equations have been developed using physics and biomechanics to describe the process of sports movement. Ramey\cite{ramey1983biomechanics} developed equations to calculate the effects of velocity-related features of the center of mass (CoM) during the flight phase(s). Seyfarth et al.\cite{seyfarth1999dynamics} modeled the supporting leg as a spring to develop equations describing the center of mass (CoM) during the take-off phase. Shang\cite{shang2022research} established mathematical models to study the influence of take-off angle, approach velocities, air density during the take-off-to-landing process. With appropriate constraints and a suitable level of abstraction, these method offer valid quantitative characterizations of specific stages within the overall movement. Nevertheless, due to the complexity of human body motions, deriving mathematical models for the entire long jump process to investigate the influence of features on final performance across all long jump phases is challenging. Furthermore, the entire movement process is non-deterministic due to the athletes’ active control, making it impossible to describe every long jump with general deterministic physical models.

As a complement, statistics have been widely adopted to investigate inter-feature relationships (e.g., correlation coefficients) and the influence of features on the target variable (e.g., linear regression). Panoutsakopoulos et al.\cite{panoutsakopoulos2021biomechanical} used simple statistics to identify the correlations between features and the effective jump distance. Kozlova et al.\cite{kozlova2020individual} employed a statistical test to assess differences between individual jumps. Dos Santos Silva et al.\cite{dos2023post} performed linear regression to study the influence of the features on jump distance. With well-established statistical tools, rigorous and robust analyses are conducted on long jump data. However, these methods struggle to capture the complex nonlinear relationships between features and the target, which restricts their capacity to reveal higher-order effects.

To overcome these issues, machine learning models that are strong at modeling nonlinear relationships have been applied to study sports-related features. Chmait et al.\cite{chmait2021artificial} summarized research that applied machine learning in sports. More specifically, U{\c{c}}ar et al.\cite{uccar2022using} applied Gradient Boosting Regression Trees (GBRT) \cite{friedman2001greedy} to predict long jump distance. XGBoost\cite{chen2016xgboost} and Random Forest\cite{breiman2001random} were used by Odong et al. \cite{odong2023introduction} to study skiing. Deep Neural Network (DNN) was used by Carvalho et al.\cite{carvalho2024swimming} and Lalwani et al.\cite{lalwani2022machine} to study swimming and volleyball, respectively. Despite the high predictive performance, these least-square-error-based models are not optimal in sports applications that are mostly interested in top-ranked performances. In the scenarios of competitive sports, more weights should be assigned to high-performing plays. Therefore, a better option is the quantile regression that aims at predicting conditional quantiles \cite{koenker2005quantile}, which could be focused on the higher-performed plays that are distributed at a higher quantile. Zhou et al.\cite{zhou2024determining} and Zhang et al.\cite{zhang2020modelling} both compared the effectiveness between quantile regression and multiple linear regression (MLR), and concluded that quantile regression could provide more effective guidance to the basketball team strategies. 
Zhao et al.\cite{zhao2024analysis} applied linear quantile regression to analyze the influence of split times on overall time in standard distance triathlons. Nevertheless, quantile-regression-based machine learning models such as quantile random forest (QRF) \cite{meinshausen2006quantile}, which combine the benefits of nonlinear-modeling-capability and quantile-focused prediction, are still rarely mentioned in sports-analysis research. 

The cost of implementing most nonlinear machine learning models, however, is the complexity to interpret compared to linear models. The most commonly applied method to the major task for machine learning models is the SHapley Additive exPlanation (SHAP) analysis \cite{lundberg2017unified}. As in Carvalho et al.\cite{carvalho2024swimming}, Lalwani et al.\cite{lalwani2022machine} and Odong et al.\cite{odong2023introduction}, feature contributions to the prediction of the target variable were evaluated by SHAP values. While SHAP values provide an overall measure of feature importance, they do not reveal how these contributions change with varying feature values. On the other hand, the Partial Dependence Plot (PDP) \cite{friedman2001greedy} and its improved variant, the Individual Conditional Expectation (ICE) plot\cite{goldstein2015peeking} provide a functional perspective on feature influence, thereby complementing SHAP analysis effectively, despite less commonly applied in sports analysis. The only example, to our best knowledge, is by Peterson\cite{peterson2018recurrent}, which analyzes the Recurrent Neural Network (RNN) with ICE on the sprint dataset.

In this paper, we adapted the commonly-used  framework in machine learning to analyze sports features. First, we incorporated quantile regression to model the non-linear function between long jump features and the distance, with a special focus on the best-performing jumps. Linear quantile regression with Lasso regularization \cite{li20081} was performed to select features and the quantile regression forests algorithm \cite{meinshausen2006quantile} was implemented to predict the long jump distance from the biomechanical features. Then, we performed feature analysis with SHAP values and ICE plots to evaluate feature importance and investigate feature interactions. The experiments were conducted on long jump datasets retrieved from the reports of the jumps from the finals of three World Championships\cite{Ref_report09, Ref_report17men, Ref_report17women, Ref_report18men, Ref_report18women}. The primary contributions of this study are as following:
\begin{itemize}
    \item We proposed to model the relationship between biomechanical features and athletic performance with QRF to enable a specific emphasis on top-tier performances from the dataset.
    \item Leveraging SHAP analysis and PDP/ICE plots, we interpreted the trained QRF model to identify key features driving elite performances and explore their interactive effects.
    \item We analyzed the contributions of biomechanical features to long jump performance in both male and female athletes, based on data from the finals of three World Championships, highlighting the critical roles of velocity and specific technical execution.
\end{itemize}

The following sections of the paper are organized as follows: The \textit{Materials and methods} section introduces the dataset, followed by a concise description of the models (quantile regression, SHAP and ICE plots) and implementation details. The \textit{Results} section presents the results of feature selection, model construction and model interpretation. Finally, the \textit{Discussion} section provides a further analysis of the findings.

\section*{Materials and methods}
This section first introduces the dataset used in this study in section 'Dataset', including an overview of the dataset, and the method used to handle the issue of missing values in the dataset. Then in section 'Methodology', the method to interpret the features in the dataset is demonstrated, which includes the models of the predictor and the interpreter and the detailed process of implementation.

\subsection*{Dataset}
% Raw data description
%\subsubsection*{Dataset description}

The dataset used in this work is drawn from five reports of the World Championships, produced by the International Amateur Athletic Federation (IAAF, the former name of World Athletics), and published on the official website of World Athletics (WorldAthletics.com). The five reports provide the biomechanical features measured from the long jump finals of women and men from three World Championships, i.e. the 2009 at Berlin, the 2017 at London and the 2018 at Birmingham (indoor). The features are from 68 jumps, among which 33 jumps are of 27 women athletes and 35 are of 25 men athletes. The total number of numeric features provided in the reports are 43. We added two extra features (height and weight of the the athletes) using online information, obtaining 45 numeric features. 
In this work, feature names are abbreviated for convenience, with explanations provided in the Supplementary Information. More detailed definitions can be found in the reports from World Athletics\cite{Ref_report09, Ref_report17men, Ref_report17women, Ref_report18men, Ref_report18women}. 
To maintain readability, explanations for each feature will also be provided in subsequent sections as needed.

The dataset is summarized in Table \ref{tab:data_overview}. The effective distance ('d\_resEffe') serving as the target variable. Its distribution is displayed in Figure \ref{fig:distance_distribution}, while the distributions of the other features can be found in the Supplementary Information. The quantiles of 'd\_resEffe' are also displayed in Figure \ref{fig:distance_distribution}. This work focuses on the explanation of the 90\% quantile. 
In this dataset, a significant number of values are missing (represented by $R_{miss}$ in Table \ref{tab:data_overview}. These missing values arise due to variations in feature sets across the three reports, accidental measurement failures, or missing online information. Consequently, the missing data fall under the categories of MAR (Missing at Random) and MCAR (Missing Completely at Random), making it suitable for imputation using algorithms (see the section Methodology).

\begin{table}%[ht]
\centering
\begin{tabular}{|l|l|l|l|l|l|l|l|}
\hline
\multirow{2}{*}{Features} & \multirow{2}{*}{Unit} & \multicolumn{3}{c|}{Men (n=35)} & \multicolumn{3}{c|}{Women (n=33)}\\
\cline{3-8}
& & Mean & S.D. & $R_{miss}(\%)$ & Mean & S.D. & $R_{miss}(\%)$ \\
\hline
d\_resOffi &m &8.09 &0.29 &0 &6.68 &0.24 &0  \\
\hline
d\_resEffe &m &8.16 &0.26 &0 &6.74 &0.24 &3 \\
\hline
d\_loss\_TO &cm &7.3 &6.4 &0 &6.5 &5.1 &3  \\
\hline
t\_step\_S3 & ms &216 &16 &23 &227 &19 &24 \\
\hline
t\_step\_S2 & ms &245 &18 &23 &244 &18 &24 \\
\hline
t\_step\_S1& ms  &195 &10 &23 &190 &16 &24 \\
\hline
t\_contact\_S3 & ms &92 &8 &23 &105 &8 &24 \\
\hline
t\_contact\_S2 & ms &113 &11 &23 &112 &13 &24 \\
\hline
t\_contact\_S1 & ms &122 &8 &23 &118 &14 &24\\
\hline
t\_flight\_S3 & ms &124 &13 &23 &123 &18 &24  \\
\hline
t\_flight\_S2 & ms &132 &17 &23 &132 &10 &24  \\
\hline
t\_flight\_S1 & ms &73 &9 &23 &73 &12 &24 \\
\hline
d\_step\_S3 &m &2.30 &0.13 &43 &2.08 &0.14 &0 \\
\hline
d\_step\_S2 & m &2.44 &0.15 &43 &2.29 &0.17 &0  \\
\hline
d\_step\_S1 & m &2.19 &0.11 &0 &2.04 &0.17 &0\\
\hline
r\_stepDiff\_S32 & \% &5.9 &4.8 &43 &10.3 &8.4 &0 \\
\hline
r\_stepDiff\_S21 & \% &-9.3 &6.2 &43 &-10.6 &6.9 &0 \\
\hline
v\_H\_S3 & m/s &10.41 &0.21 &43 &9.29 &0.27 &0  \\
\hline
v\_H\_S2 & m/s &10.37 &0.28 &43 &9.30 &0.27 &0  \\
\hline
v\_H\_S1 & m/s &9.76 &0.41 &0 &8.87 &0.39 &0  \\
\hline
v\_H\_TO & m/s &8.68 &0.45 &0 &7.93 &0.42 &0 \\
\hline
v\_V\_TO & m/s &3.68 &0.29 &0 &3.15 &0.31 &0  \\
\hline
t\_TDO & s &0.12 &0.01 &77 &0.12 &0.01 &76  \\
\hline
v\_HDiff\_TDO & m/s &-1.59 &0.50 &0 &-1.45 &0.34 &0 \\
\hline
v\_TO & m/s &9.43 &0.37 &0 &8.54 &0.38 &0  \\
\hline
a\_TO & ° &23.0 &2.3 &0 &21.7 &2.4 &0  \\
\hline
h\_CMLower & cm &3.3 &1.6 &23 &3.2 &1.9 &24  \\
\hline
a\_body\_TD & ° &-35.5 &2.1 &23 &-36.1 &1.8 &24  \\
\hline
a\_body\_TO & °  &19.9 &3.9 &0 &19.5 &4.8 &0 \\
\hline
a\_trunk\_TD & ° &-5.6 &4.4 &57 &-7.4 &4.6 &61 \\
\hline
a\_trunk\_TO & ° &1.9 &6.5 &0 &3.3 &6.1 &0  \\
\hline
a\_trunkRot\_TDO & ° &10.1 &2.8 &77 &5.6 &4.5 &76 \\
\hline
a\_thigh\_TO & ° &-13.7 &9.0 &0 &-10.7 &8.9 &0 \\
\hline
w\_thigh\_TDO & °/s &598 &141 &0 &616 &127 &0 \\
 \hline
a\_knee\_TD & ° &169.3 &5.6 &23 &167.9 &6.3 &24 \\
\hline
a\_kneeMin\_TDO & ° &138.8 &8.8 &0 &138.2 &5.8 &0 \\
\hline
a\_kneeRange\_TDO & ° &31.0 &9.2 &23 &29.0 &6.6 &24  \\
\hline
w\_knee\_TDO & °/s &-504 &143 &23 &-473 &115 &24  \\
\hline
a\_hip\_LD & ° &94.0 &19.0 &37 &88.4 &14.0 &36  \\
\hline
a\_knee\_LD & ° &134.7 &12.1 &37 &138.1 &14.7 &36  \\
\hline
a\_trunk\_LD & ° &28.4 &40.2 &37 &31.3 &35.3 &36  \\
\hline
d\_loss\_LD & m &0.06 &0.10 &37 &0.05 &0.07 &36 \\
\hline
d\_LD & m &0.62 &0.13 &37 &0.55 &0.08 &36 \\
\hline
Height & m &1.85 &0.06 &0 &1.73 &0.06 &3 \\
\hline
Weight & kg &75 &7 &0 &62 &6 &6 \\
\hline
\end{tabular}
\caption{\label{tab:data_overview}Dataset overview. The basic statistics of each feature (including the target variable 'd\_resEffe') of each gender are provided. The percentages of missing values are also provided in the column $R_{miss}$.}
\end{table}

\begin{figure}[ht]
    %\centering  
    % First plot
    \begin{subfigure}{0.49\textwidth}
        \centering
        \includegraphics[width=\textwidth]{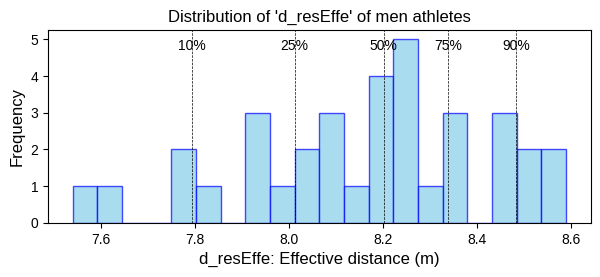}
        \caption{Distribution of the 'd\_resEffe' of the men athletes' jumps.}
        \label{fig:dist_hist_men}
    \end{subfigure}
    %\vspace{1cm} % Space between the two plots
    % Second plot
    \begin{subfigure}{0.49\textwidth}
        \centering
        \includegraphics[width=\textwidth]{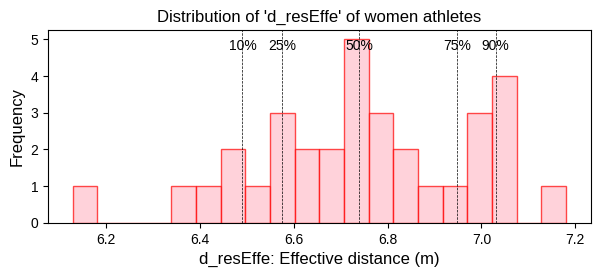}
        \caption{Distribution of the 'd\_resEffe' of the women athletes' jumps.}
        \label{fig:dist_hist_women}
    \end{subfigure}
    \caption{Distribution of the target variable 'd\_resEffe' (effective distance) of each gender. Typical quantiles are marked with vertical dash lines. }
    \label{fig:distance_distribution}
\end{figure}

\subsection*{Methodology}
In this section, the basics of the methods and the experimental details are introduced. Male and female samples were analyzed separately for two main reasons. First, there are significant differences in the distributions of both biomechanical features and target variables between the two genders. Training separate models allows for better capture of intra-gender variation compared to using a single model on the combined dataset (see the Supplementary Information for details). Second, since our focus is on top performances and male athletes generally exhibit much higher performance levels than female athletes, combining both genders would obscure the analysis specific to female athletes. The experimental steps in the pipeline include feature imputation, feature selection with Lasso regularized linear quantile regression\cite{li20081}, model training through quantile random forests\cite{meinshausen2006quantile}, model interpretation with SHAP analysis\cite{lundberg2017unified} and PDP and ICE plots\cite{goldstein2015peeking}.

\subsubsection*{Feature imputation}
Four imputing methods to estimate missing features were experimented: mean value imputation, KNN imputation\cite{troyanskaya2001missing, batista2002study}, Bayesian-Ridge-based iterative imputation\cite{mackay1992bayesian, tipping2001sparse} and Random Forest-based iterative imputation\cite{breiman2001random, tang2017random}. They were performed on samples of men and women, respectively. The selection of the imputation methods was done with nested cross-validation, with 4 folders of outer loops and 3 folders of inner loops. To guarantee a good estimation of missing values across all the samples, rather than only the highest performed ones, we used Random Forest\cite{breiman2001random} as the model and with mean squared error(MSE), root mean squared error (RMSE) and R-squared ($R^2$) as the metrics, instead of their quantile regression variant.

\subsubsection*{Feature selection with quantile Lasso}
There are abundant redundancies in the feature set, which leads to the necessity of performing feature selection before training a predictor model. The redundancies arise from the definitions of the features, where some are mathematically derived from others, as well as from the underlying physical relationships in the long jump process. 
To drop redundant features while preserving the most useful information, we select features correspond to the optimal model prediction performance.
We implemented the quantile linear regression with Lasso regularization for feature selection following Li et al.\cite{li20081} The objective of Lasso-regularized quantile linear regression is to optimize the following function:
\begin{gather}
    \min\limits_{\beta_0,\bm{\beta}} \sum_{i=1}^n\rho_{\tau}(y_i - \beta_0 - \bm{\beta}^T\bm{x_i})\\
    \bm{\beta} = [\beta_1, ..., \beta_p]^T \\
    \text{subject to } |\beta_1| + \cdots + |\beta_p| \leq s
    \label{eq:lasso_reg}
\end{gather}
where $\bm{x_i} \in \mathbb{R}^{p}$ represents the $i$-th sample ($1 \leq i \leq n$) with p features, $y_i \in \mathbb{R}$ is the corresponding target, $\beta_0$ and $\bm{\beta}$ are the regression coefficients and $s$ serves as a hyperparameter to control the strength of regularization, with smaller values of $s$ indicating stronger regularization. The pinball loss function, commonly used for estimating conditional quantiles~\cite{steinwart2011estimating}, assigns asymmetric weights to overestimation and underestimation errors based on the specified quantile level. We represent the pinball loss as $\rho_\tau(\cdot)$, which is defined as:
\begin{equation}
    \rho_{\tau}(u) =
    \begin{cases} 
    \tau \cdot u & \text{if } u > 0, \\
    -(1-\tau)\cdot u & \text{if } u \leq 0.
    \end{cases}
    \label{eq:pinball}
\end{equation}

%The implementation is from \REF… 
Three-fold cross-validation was performed ten times, each with a different random seed. The average coefficients and pinball loss were recorded for selecting the best-performing feature sets.

\subsubsection*{Quantile regression forests}
The predictor model was built with quantile regression forests, a generalization of random forests that could infer conditional quantiles\cite{meinshausen2006quantile}. The implementation is carried out with the Python package \textit{quantile-forest} (version 1.3.11). When tuning hyperparameters, a large number of trees were set to introduce sufficient randomness and thus improve the robustness of the results. Furthermore, the model was evaluated with K-fold cross-validation, with K set to 4 in the experiments. As mentioned previously, the quantiles for both men and women were set to 90\%, and the evaluation was based on pinball loss (see Equation \ref{eq:pinball})

\subsubsection*{SHAP analysis}
%(A brief introduction, and then the details to implement it)
SHAP (SHapley Additive exPlanations) analysis is based on game theory and Shapley values to interpret feature contributions to the output of a given machine learning model\cite{lundberg2017unified}. For each prediction, a SHAP value is assigned to each feature. The mean absolute value of SHAP values of a particular feature aggregates the contributions of the feature across all the samples, providing a global explanation of the feature importance. We applied the SHAP analysis using the Python package \textit{shap} (version 0.45.1).

\subsubsection*{PDP and ICE plots}
PDP (Partial Dependence Plots) and ICE (Individual Conditional Expectation)  plots interpret machine learning models by visualizing marginal feature effects. According to Goldstein\cite{goldstein2015peeking}, consider a regressor $f:\mathbb{R}^p\mapsto\mathbb{R}$ and an estimation of $f$ using quantile random forests, noted as $\hat{f}^{QRF}$. For $S \subset \{1,...,p\}$ and $C$ being the complement set of $S$, the partial dependence function is calculated by
\begin{equation}
    f_S = \mathbb{E}_{\bm{x}_C}[f(\bm{x}_S,\bm{x}_C)] = \int f(\bm{x}_S,\bm{x}_C) dP(\bm{x}_C)) \sim \frac{1}{N}\sum_{i=1}^{N}\hat{f}^{QRF}(\bm{x}_S,\bm{x}^{(i)}_{C})
\end{equation}
where $\bm{x}_{S}$ denotes the values of features in set $S$ and $\bm{x}^{(i)}_{C}$ represents the values of features in set $C$ for sample $i$, respectively, and $N$ is the number of samples. PDP shows the average numerical effects of one or a set of feature(s) on the model output. ICE plot provides more details by plotting each individual function:
\begin{equation}
    \hat{f}^{(i)}_S=\hat{f}^{QRF}(\bm{x}_S,\bm{x}^{(i)}_{C})
\end{equation}

In this work, $f$ is treated as the predictive model that estimates long jump distance based on the biomechanical features. By designating a specific feature as $\bm{x}_{S}$ and grouping the remaining features as $\bm{x}_{C}$, we utilized ICE plots to visualize the quantitative influence of $\bm{x}_{S}$. The implementation was based on the \textit{scikit-learn} library\cite{pedregosa2011scikit} (version 1.3.0).

\section*{Results}

The results are divided into the model development section and the model interpretation section. In the first section, details of feature imputation, feature selection and model training are provided. In the subsequent section, analysis through SHAP values and ICE plots are presented.

\subsection*{Model development}

\subsubsection*{Feature imputation}
The results are shown in Table \ref{tab:imputation}. The MSE, RMSE and $R^2$ metrics were obtained with the process introduced in Methodology. According to the metrics, Random Forest iterative imputation achieved the lowest error for men's samples and KNN imputation achieved the lowest error for women's samples. Data imputed by these two methods for corresponding gender were used in the following experiments. 

\begin{table}[ht]
    \centering
    \begin{tabular}{|l|c c c| c c c|}
        \hline
        \multirow{2}{*}{Method} & \multicolumn{3}{c|}{Men} & \multicolumn{3}{c|}{Women}\\
        \cline{2-7}
        & MSE$(\downarrow)$ & RMSE$(\downarrow)$ & $R^2$$(\uparrow)$ & MSE$(\downarrow)$ & RMSE$(\downarrow)$ & $R^2$$(\uparrow)$\\ 
        \hline
        Mean & 0.0493 & 0.222 & 0.262 & 0.0452 & 0.213 & 0.324 \\
        \hline
        KNN & 0.0435 & 0.208 & 0.350 & \textbf{0.0421} & \textbf{0.205} & \textbf{0.370}\\
        \hline
        Bayesian & 0.0484 & 0.220 & 0.277 & 0.0445 & 0.211 & 0.334 \\
        \hline
        RF & \textbf{0.0432} & \textbf{0.208} & \textbf{0.354} & 0.0436 & 0.209 & 0.349 \\
        \hline
    \end{tabular}
    \caption{\label{tab:imputation} Evaluation of four imputation methods using MSE, RMSE and $R^2$ for both genders, with the best metric for each gender highlighted in \textbf{bold}. The four methods are the mean imputation, the KNN imputation, the Bayesian-Ridge-based iterative imputation and the Random-Forest-based iterative imputation.}
\end{table}

\begin{figure}[ht]
    %\centering  
    % First plot
    \begin{subfigure}{0.49\textwidth}
        \centering
        \includegraphics[width=\textwidth]{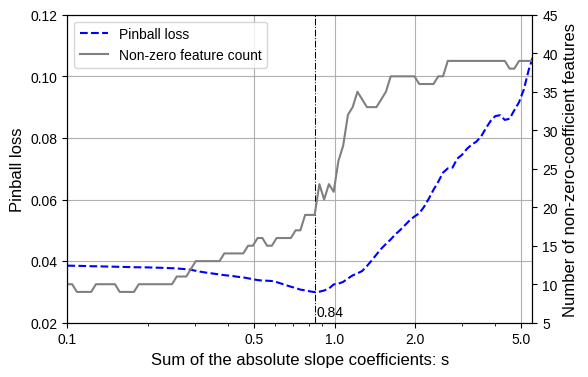}
        \caption{Results of the men athletes' jumps.}
        \label{fig:feature_selection_men}
    \end{subfigure}
    %\vspace{1cm} % Space between the two plots
    % Second plot
    \begin{subfigure}{0.49\textwidth}
        \centering
        \includegraphics[width=\textwidth]{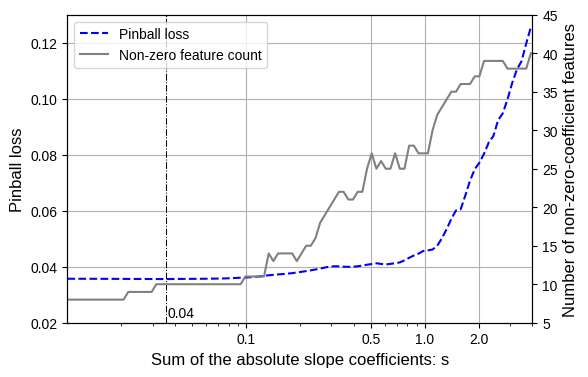}
        \caption{Results of the women athletes' jumps.}
        \label{fig:feature_selection_women}
    \end{subfigure}
    \caption{Results of feature selection with Lasso regularization. The changes of pinball loss of model predictions against the sum of the absolute slope coefficients $s$ are in blue dash curves, while the non-zero feature counts against $s$ are in gray solid curves. The results are displayed by genders: (\textbf{a}) men athletes' jumps and (\textbf{b}) women athletes' jumps. }
    \label{fig:feature_selection}
\end{figure}

\subsubsection*{Feature selection}

The results of the feature selection by Lasso-regularized linear quantile regression are summarized in Figure \ref{fig:feature_selection}. Figure \ref{fig:feature_selection_men} shows the results on the men's jumps and Figure \ref{fig:feature_selection_women} shows the results on the women's jumps. Each plot shows the variation of model performance (blue dashed curves) and non-zero feature count (gray solid curves) with respect to the sum of the absolute slope coefficient $s$. $s$ is the regularization strength of the Lasso penalty: the smaller $s$ is, the stronger regularization will be, and the less non-zero features tend to be obtained (see Equation \ref{eq:lasso_reg}). The value of $s$ was selected to ensure model predictions with the minimum pinball loss. In each plot, the minimum-pinball-loss $s$ value was provided with a vertical dashed line. For men and women atheltes' samples, $s$ were selected as 0.84 and 0.04, respectively. The selected features were as follows: 
\begin{itemize}
    \item Men athletes' jumps (19 features): 'v\_H\_S1', 'a\_knee\_TD', 't\_contact\_S2', 'd\_loss\_LD', 'h\_CMLower', 'v\_TO', 'a\_knee\_LD', 'Height', 't\_flight\_S1', 'Weight', 't\_flight\_S2', 'v\_H\_S3', 'a\_trunk\_TO', 'v\_V\_TO', 'a\_kneeRange\_TDO', 'r\_stepDiff\_S21', 't\_flight\_S3', 'w\_thigh\_TDO', 'v\_H\_S2'.
    \item Women athletes' jumps (10 features): 'v\_H\_S2', 'd\_LD', 'r\_stepDiff\_S32', 'a\_knee\_LD', 't\_flight\_S2', 'r\_stepDiff\_S21', 'a\_thigh\_TO', 'v\_H\_S3', 't\_flight\_S1', 'v\_H\_S1'.
\end{itemize}

\subsubsection*{Model training}
% Model fitting results

Given the distinct performance distributions between genders (see Figure \ref{fig:distance_distribution}), separate quantile random forest models were trained for male and female athletes' samples. The hyperparameters ('n\_estimators', 'max\_depth' and 'max\_features') of the quantile random forest were tuned through K-fold cross validation on the full data of men (or women) samples, with K being 4. The mean pinball loss was used as the evaluation metric. For male athletes, the optimal parameters we obtained are 100 tree estimators, a maximum depth of 3, and a maximum of 6 features, while for female athletes, the optimal parameters are 100 tree estimators, a maximum depth of 3, and a maximum of 1 features. The best metrics obtained during the cross-validation process from men and women samples were 0.0287 and 0.0333, respectively. Then with the tuned hyperparameters, the model was trained on the full data of men (or women) samples. Limited by dataset size, we did not take samples to build a test set.

\begin{figure}[ht]
    %\centering  
    % First plot
    \begin{subfigure}{0.5\textwidth}
        \centering
        \includegraphics[width=\textwidth]{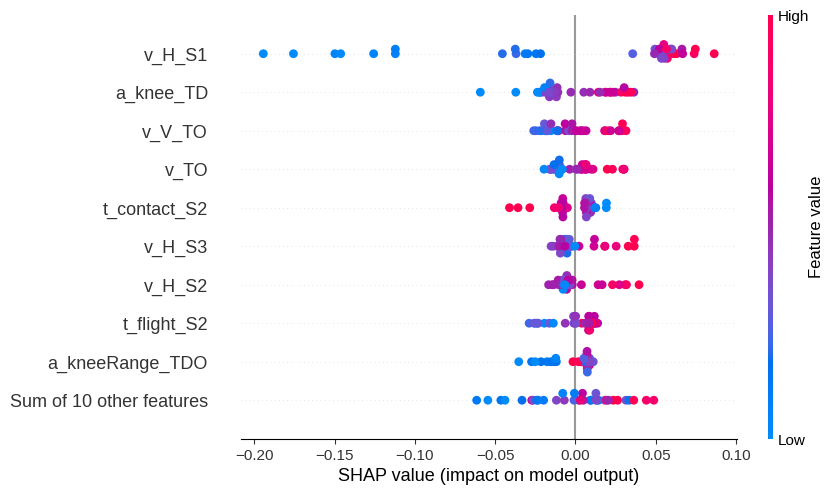}
        \caption{SHAP beeswarm plot on the men athletes' jumps.}
        \label{fig:shap_beeswarm_men}
    \end{subfigure}
    %\vspace{1cm} % Space between the two plots
    % Second plot
    \begin{subfigure}{0.45\textwidth}
        \centering
        \includegraphics[width=\textwidth]{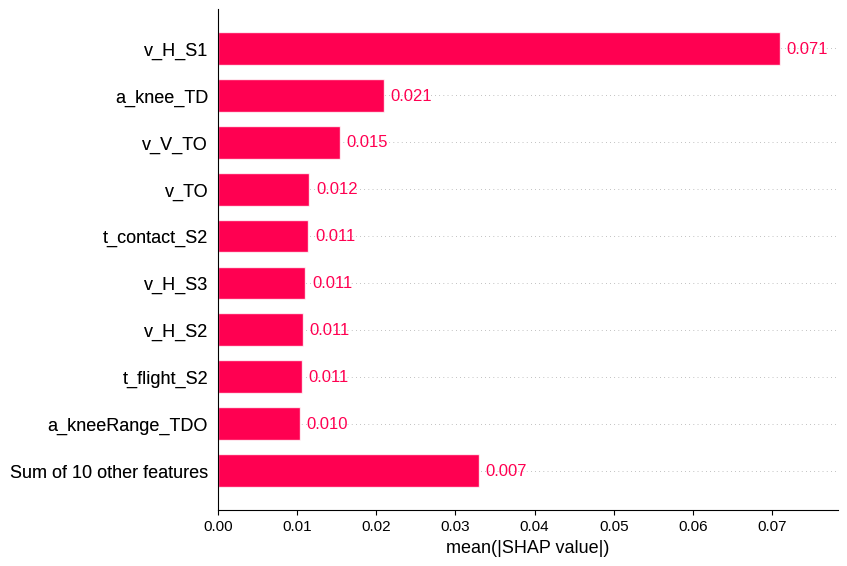}
        \caption{SHAP bar plot on the men athletes' jumps.}
        \label{fig:shap_bar_men}
    \end{subfigure}

    \begin{subfigure}{0.5\textwidth}
        \centering
        \includegraphics[width=\textwidth]{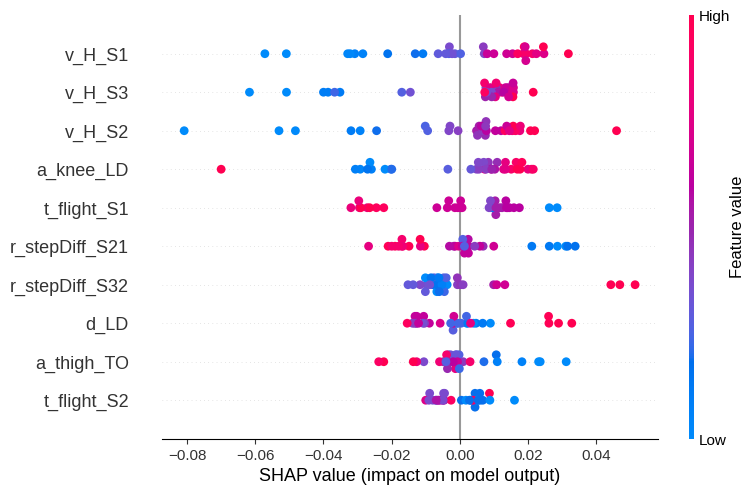}
        \caption{SHAP beeswarm plot on the women athletes' jumps.}
        \label{fig:shap_beeswarm_women}
    \end{subfigure}
    %\vspace{1cm} % Space between the two plots
    % Second plot
    \begin{subfigure}{0.45\textwidth}
        \centering
        \includegraphics[width=\textwidth]{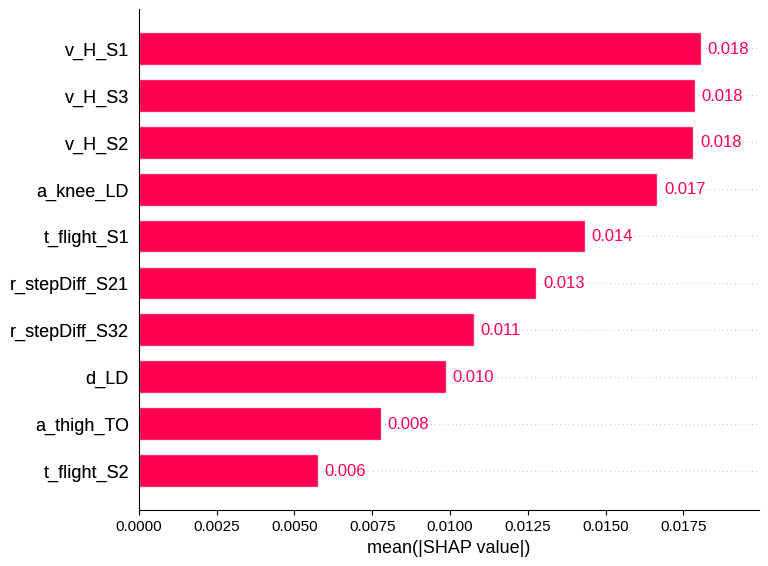}
        \caption{SHAP bar plot on the women athletes' jumps.}
        \label{fig:shap_bar_women}
    \end{subfigure}
    \caption{Global SHAP analysis on the men and women athletes' jumps, respectively. (\textbf{a}, \textbf{b}) on the men athletes' jumps, (\textbf{c}, \textbf{d}) on the women athletes' jumps.}
    \label{fig:shap_global}
\end{figure}

\subsection*{Model interpretation}
% SHAP analysis results

\subsubsection*{Global SHAP analysis}

The global SHAP results are presented in Figure \ref{fig:shap_global} where the features are arranged along the vertical axis in descending order of their shap values. The beeswarm plots (Figure \ref{fig:shap_beeswarm_men} and \ref{fig:shap_beeswarm_women}) provide a detailed visualization, with each sample represented as a point for each feature. The horizontal position of each point corresponds to the SHAP value of the feature for the respective sample, reflecting the feature's contribution to the model's prediction. In addition, the color of the points indicates the values of the features. The bar plots (\ref{fig:shap_bar_men} and \ref{fig:shap_bar_women}) show the mean absolute SHAP values for each feature in the corresponding beeswarm plot. 

For men's jumps, the velocity-related features dominate the contributions. The most important feature is ‘v\_H\_S1’ (the mean horizontal velocity during the last step), with the largest mean absolute SHAP value, being 0.071. The ‘beeswarm’ plot shows that ‘v\_H\_S1’ has a positive contribution, i.e. the larger this velocity is, the longer the distance is. The third to the eighth most important features are also related to the horizontal velocities, which are ‘v\_V\_TO’ (vertical velocity at take-off), ‘v\_TO’ (the velocity magnitude at take-off), ‘t\_contact\_S2’ (the contact time of the second last step),  ‘v\_H\_S3’ (the mean horizontal velocity of the third last step), ‘v\_H\_S2’ (the mean horizontal velocity of the second last step) and 't\_flight\_S2' (the flight time of the second last step). ‘v\_H\_TO’, ‘v\_V\_TO’ and ‘v\_TO’ reveal the relationship between the magnitude and the angle of the take-off velocity. According to the model proposed by Linthorne et al. ~\cite{linthorne2005optimum}, the optimal take-off angle is a function of the take-off velocity. Furthermore, the study of Dorn et al.~\cite{dorn2012muscular} demonstrated the influence of contact time and flight time on horizontal velocity during sprinting.  

Another group of the highly-ranked features for men athletes' jumps is related to the process of transferring horizontal momentum to vertical, i.e. the second and the ninth most important features, which are ‘a\_knee\_TD’ (the knee angle when foot touchdown on the board) and ‘a\_kneeRange\_TDO’ (the range of the knee angle from touchdown on the board to the minimum before take-off), respectively. Note that the knee angle is defined as 180° when the leg is stretched straight. Thus, the positive contribution of ‘a\_knee\_TD’ in the ‘beeswarm’ plot indicates that the more the support leg stretches when the athlete touches down on the board, the longer distance the jump is likely to get. On the other hand, together with ‘a\_kneeRange\_TDO’, the two knee-related features describe the leg bending process before take-off, which is related to the process of transferring horizontal momentum to vertical. The process was studied in literature \cite{seyfarth1999dynamics, seyfarth2000optimum, muraki2005athletics} that modeled the supporting leg as a linear spring. 

For women's jumps, velocity-related features also play dominant roles. ‘v\_H\_S1’, ‘v\_H\_S3’ and ‘v\_H\_S2’ are the most important features, whose mean absolute SHAP values are all around 0.018. In addition, features 't\_flight\_S1', 't\_flight\_S2' are also related to velocity. Unlike men's jumps, two critical groups of features are evident for women athletes. The first pertains to approach step adjustments, including 'r\_stepDiff\_S21' and 'r\_stepDiff\_S32' (the step length change percentage from second last step to the last step and from the third last step to the last step, respectively). The second group relates to the landing pose, encompassing 'a\_knee\_LD' (the knee angle of the landing leg when landing in the sand pit) and 'd\_LD' (the horizontal distance between the center of mass and the landing foot when landing in the sand pit). These findings highlight that approach techniques and landing techniques are critical for World-Championship-level women athletes to achieve high performance.

\begin{figure}[ht]
    %\centering  
    % First plot
    \begin{subfigure}{0.5\textwidth}
        \centering
        \includegraphics[width=\textwidth]{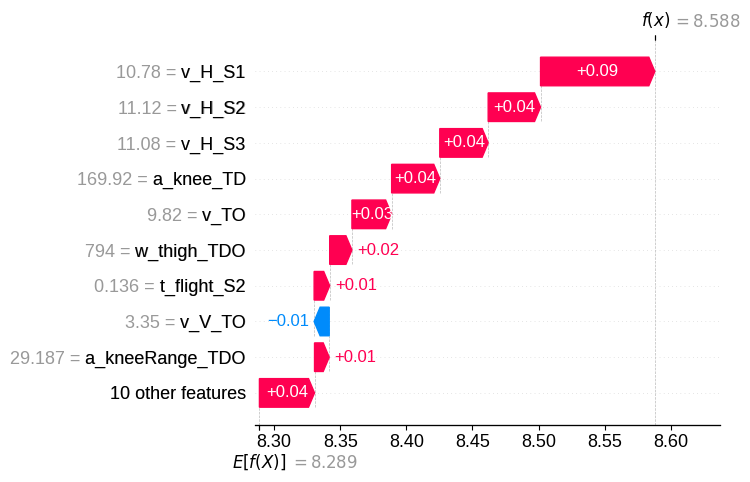}
        \caption{The best men athlete's jump: 8.59m.}
        \label{fig:shap_best_men}
    \end{subfigure}
    \begin{subfigure}{0.45\textwidth}
        \centering
        \includegraphics[width=\textwidth]{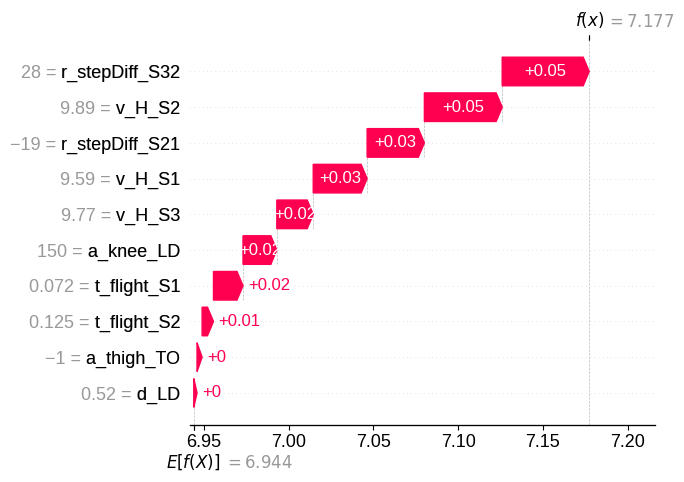}
        \caption{The best women athlete's jump: 7.18m.}
        \label{fig:shap_best_women}
    \end{subfigure}
    \caption{Individual SHAP results of the best jumps of the men and women athletes' jumps, respectively.}
    \label{fig:shap_individual}
\end{figure}

\subsubsection*{Individual SHAP analysis}

The SHAP values of the best jumps for men and women are presented in Figure \ref{fig:shap_individual}. In each plot, the contributions of each feature to the model prediction are provided, which add up to the final prediction, given by $f(x)$. The final predictions of the performances (indicated by $f(x)$ in the plots) are very close to the ground truths (given in the captions of each plot), which shows the good predictive performance of the trained model.

For the best jump of the men athletes (see Figure \ref{fig:shap_best_men}), the most important feature is 'v\_H\_S1', which is the same as the results of the global SHAP. The velocities of the second and the third last step ('v\_H\_S2' and 'v\_H\_S3') ranked the second and third, respectively. It is noticeable that the three velocities of this jump are 10.78m/s, 11.12m/s and 11.08m/s, which are all significantly higher than the average velocities (9.76m/s, 10.37m/s and 10.41m/s, respectively) and even the second highest velocities (10.44m/s, 10.82m/s and 10.59m/s, respectively). Following them, 'a\_knee\_TD' ranked the fourth and 'a\_kneeRange\_TDO' ranked the nineth, which indicates that the knee movement is still important, but dominated by the outstanding velocities. %Another interesting point is that the 'v\_V\_TO' has a negative contribution to the performance. This value of the jump is 3.35m/s, which is lower than the average value, 3.68m/s. 

For the best jump of the women atheltes (see Figure \ref{fig:pdp_1_women}), the velocity-related features are less important compared to the results for the best men's jump. The three velocities ('v\_H\_S1', 'v\_H\_S2' and 'v\_H\_S3') of the best women's jump are 9.59m/d, 9.89m/s and 9.77m/s, which are the best among all the women's jumps. Compared with the second highest velocities of the three features (not from the same jump), i.e. 9.52m/s, 9.68m/s and 9.76m/s, the velocities of the third last step are very close, but differ much more for the second last and the last step. Meanwhile, the first and third important features for this jump are 'r\_stepDiff\_S21' and 'r\_stepDiff\_S32', which are both the largest in magnitude among the women's samples. It suggests that a good approach technique has been applied to the last few steps of the best women's jump, which help to achieve the longest effective distance. In addition, the positive 'r\_stepDiff\_S32' (28\%) and negative 'r\_stepDiff\_S21' (-19\%) follow the commonly used 'larger penultimate - shorter last step' technique as mentioned by Panoutsakopoulos et al. \cite{panoutsakopoulos2021biomechanical, panoutsakopoulos2017gender} % which is investigated in later parts

\begin{figure}[ht]
    %\centering  
    % First plot
    \begin{subfigure}{0.33\textwidth}
        \centering
        \includegraphics[width=\textwidth]{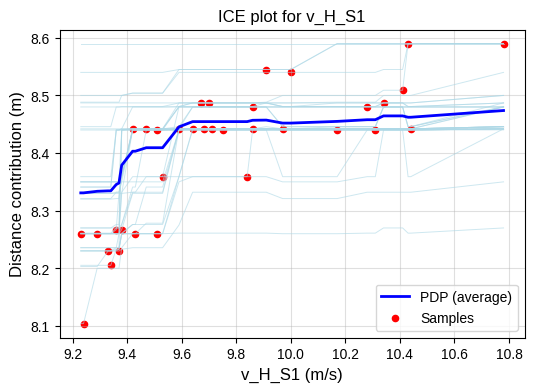}
        \caption{ICE plot of 'v\_H\_S1'.}
        \label{fig:ice_1_men}
    \end{subfigure}
    \begin{subfigure}{0.33\textwidth}
        \centering
        \includegraphics[width=\textwidth]{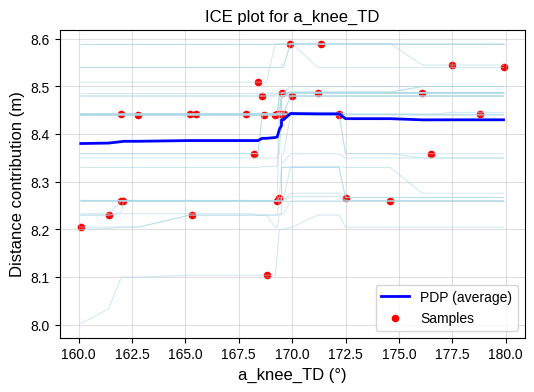}
        \caption{ICE plot of 'a\_knee\_TD'.}
        \label{fig:ice_2_men}
    \end{subfigure}
    \begin{subfigure}{0.33\textwidth}
        \centering
        \includegraphics[width=\textwidth]{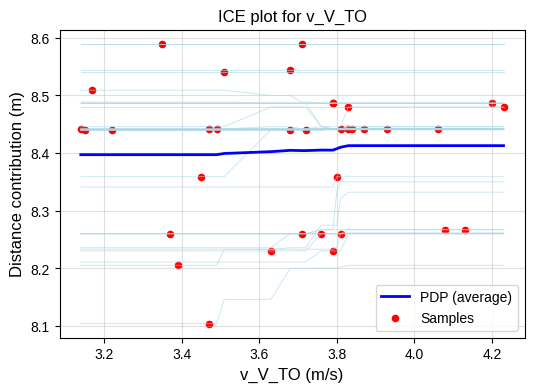}
        \caption{ICE plot of 'v\_V\_TO'.}
        \label{fig:ice_3_men}
    \end{subfigure}
    %\vspace{1cm} % Space between the two plots
    % Second plot
    \begin{subfigure}{0.33\textwidth}
        \centering
        \includegraphics[width=\textwidth]{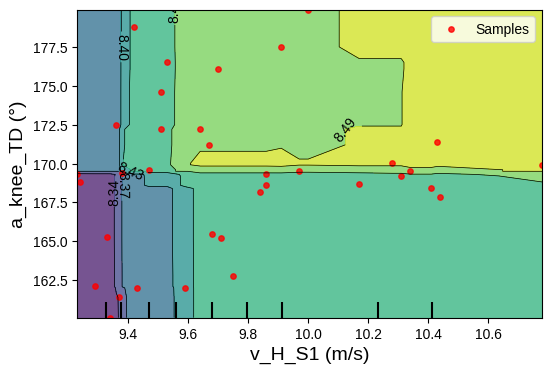}
        \caption{PDP of 'v\_H\_S1' and 'a\_knee\_TD'.}
        \label{fig:pdp_1_men}
    \end{subfigure}
    \begin{subfigure}{0.33\textwidth}
        \centering
        \includegraphics[width=\textwidth]{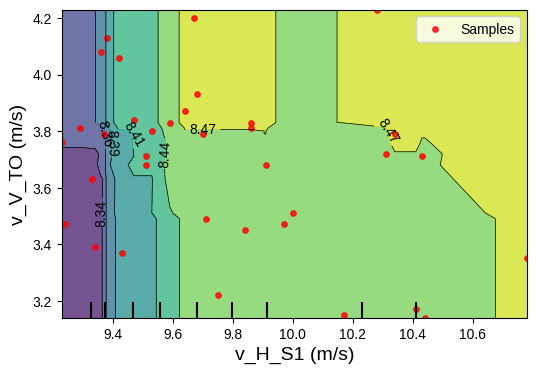}
        \caption{PDP of 'v\_H\_S1' and 'v\_V\_TO'.}
        \label{fig:pdp_2_men}
    \end{subfigure}
    \begin{subfigure}{0.33\textwidth}
        \centering
        \includegraphics[width=\textwidth]{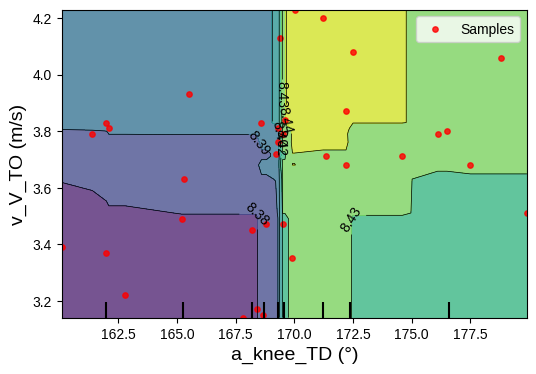}
        \caption{PDP of 'a\_knee\_TD' and 'v\_V\_TO'.}
        \label{fig:pdp_3_men}
    \end{subfigure}
    \caption{ICE plots and 2D PDP of the top ranked features for the male athletes' jumps given by SHAP analysis. The analyzed features include 'v\_H\_S1' (the mean horizontal velocity of the last step), 'a\_knee\_TD' (the knee angle of the supporting leg at touchdown) and 'v\_V\_TO' (the vertical velocity at take-off).}
    \label{fig:ice_pdp_men}
\end{figure}

\subsubsection*{ICE plot analysis}

\begin{figure}[ht]
    %\centering  
    % First plot
    \begin{subfigure}{0.33\textwidth}
        \centering
        \includegraphics[width=\textwidth]{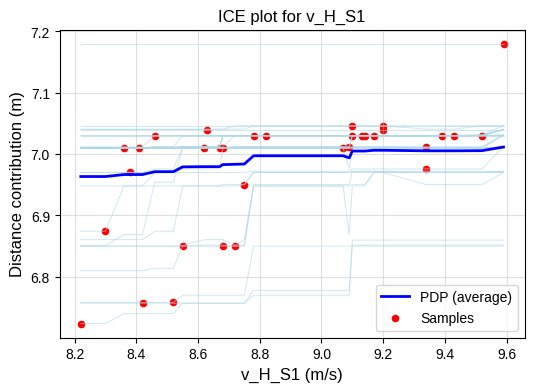}
        \caption{ICE plot of 'v\_H\_S1'.}
        \label{fig:ice_1_women}
    \end{subfigure}
    \begin{subfigure}{0.33\textwidth}
        \centering
        \includegraphics[width=\textwidth]{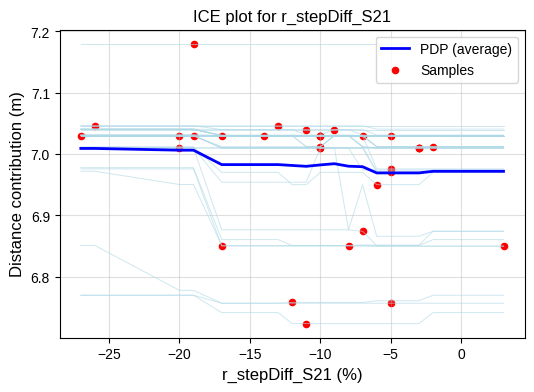}
        \caption{ICE plot of 'r\_stepDiff\_S21'.}
        \label{fig:ice_2_women}
    \end{subfigure}
    \begin{subfigure}{0.33\textwidth}
        \centering
        \includegraphics[width=\textwidth]{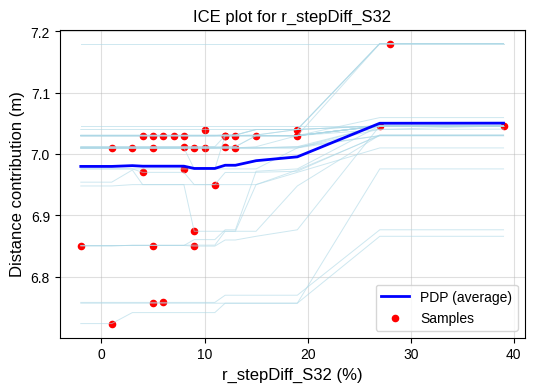}
        \caption{ICE plot of 'r\_stepDiff\_S32'.}
        \label{fig:ice_3_women}
    \end{subfigure}
    %\vspace{1cm} % Space between the two plots
    % Second plot
    \begin{subfigure}{0.33\textwidth}
        \centering
        \includegraphics[width=\textwidth]{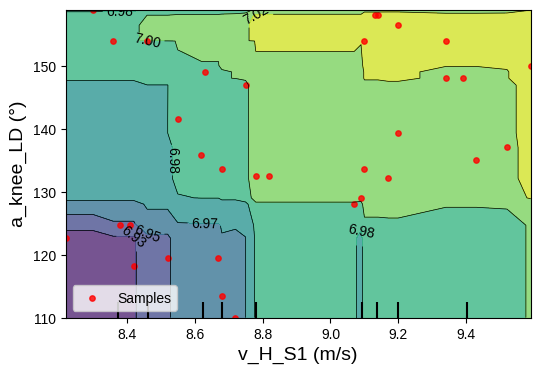}
        \caption{PDP of 'v\_H\_S1' and 'a\_knee\_LD'.}
        \label{fig:pdp_1_women}
    \end{subfigure}
    \begin{subfigure}{0.33\textwidth}
        \centering
        \includegraphics[width=\textwidth]{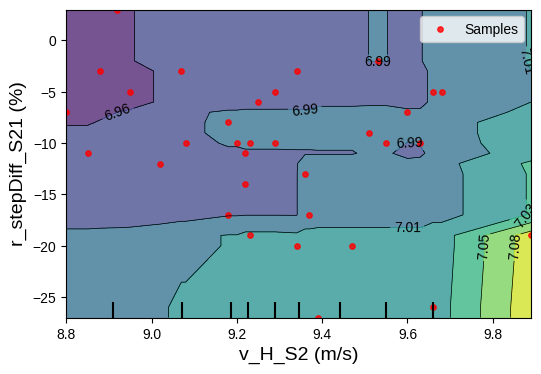}
        \caption{PDP of 'v\_H\_S2' and 'r\_stepDiff\_S21'.}
        \label{fig:pdp_2_women}
    \end{subfigure}
    \begin{subfigure}{0.33\textwidth}
        \centering
        \includegraphics[width=\textwidth]{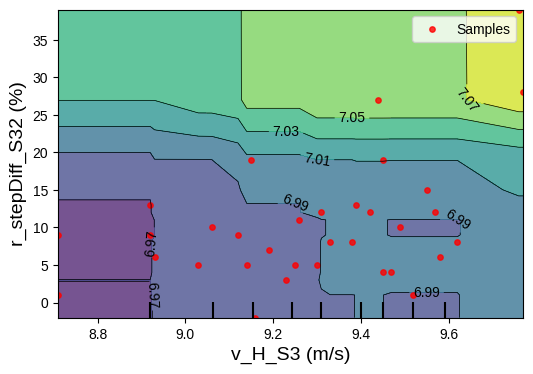}
        \caption{PDP of 'v\_H\_S3' and 'r\_stepDiff\_S32'.}
        \label{fig:pdp_3_women}
    \end{subfigure}
    \caption{ICE plots and 2D PDP of the top ranked features for the female athletes' jumps given by SHAP analysis. We analyze for ICE  the features (a) 'v\_H\_S1' the mean horizontal velocity of the last step, (b) 'v\_H\_S2' the second last step , (c) 'v\_H\_S3' the third last step. We analyze for PDP (d) 'a\_knee\_LD' the knee angle of the landing leg when landing in the sand pit, (e) 'r\_stepDiff\_S21' the step length change percentage from second last step to the last step and (f) 'r\_stepDiff\_S32' the step length change percentage from the third last step to the last step.}
    \label{fig:ice_pdp_women}
\end{figure}

The ICE plots of the men and women athletes' jumps are displayed in Figure \ref{fig:ice_pdp_men} and \ref{fig:ice_pdp_women}, respectively. Each figure contains three ICE plots and three 2D PDP. In each ICE plot, the detailed influence of a feature is illustrated by the ICE curves and the PDP curve which is the average of all the ICE curves. Each 2D PDP shows the interactive effects of two features on the target variable which is displayed by contours. Due to the limited number of samples, there are extrapolation regions in the plots. To better visualize the extrapolated regions, the samples are also plotted with red points. In this part, some of the most important and representative features were selected for analysis.

Previous results indicate 'v\_H\_S1' as the most important feature for men athletes, but without much details about its numeric impacts. For more insights, in Figure \ref{fig:ice_1_men}, it is shown that there are two types of impacts according to the values of 'v\_H\_S1'. When it is below 9.6m/s, the influence of 'v\_H\_S1' is stronger than when the velocity is above 9.6m/s, according to the steepness of the PDP and ICE curves. Moreover, the dispersion of the sample distribution at each value of 'v\_H\_S1' suggests the influence of the other features on the performance. Figure \ref{fig:pdp_1_men} and \ref{fig:pdp_2_men} illustrate the co-influence of 'v\_H\_S1' with 'a\_knee\_TD' and 'v\_V\_TO', respectively. In Figure \ref{fig:pdp_1_men}, there are two apparent thresholds. The first is 9.6m/s for 'v\_H\_S1'. The other is around 169° for 'a\_knee\_TD'. The performance improved significantly when the values of the two features are above these two thresholds. In Figure \ref{fig:pdp_2_men}, 'v\_V\_TO' has a minor positive influence compared with 'v\_H\_S1'.  These interactions can further explain the results in Figure \ref{fig:ice_2_men} and \ref{fig:ice_3_men}. In Figure \ref{fig:ice_2_men}, there is a stepwise increase at around 169° of 'a\_knee\_TD', and the large variances of the distance at specific values of 'a\_knee\_TD' suggest that the effective distance is also strongly influenced by some other features, e.g. 'v\_H\_S1'. Figure \ref{fig:ice_3_men} shows the interaction between 'a\_knee\_TD' and 'v\_V\_TO'. First of all, according to the sample distribution, 'a\_knee\_TD' shows a positive linear correlation with 'v\_V\_TO'. This correlation suggests that when 'a\_knee\_TD' is too small, 'v\_V\_TO' cannot be too large. On the plot, it corresponds to the empty region at the left-above corner. Secondly, for a given value of 'a\_knee\_TD', a larger 'v\_V\_TO' is favored for a better performance. Finally, the effect of threshold is more significant than in Figure \ref{fig:pdp_1_men}. Regardless of the value of 'v\_V\_TO', when 'a\_knee\_TD' exceeds 169°, the effective distance is consistently greater than when 'a\_knee\_TD' is below this threshold.

Similarly, 'v\_H\_S1' is also the most important feature for female athletes' jumps. Figure \ref{fig:ice_1_women} is the ICE plot of 'v\_H\_S1' of the women's samples. The tendency of PDP is similar to that of the men's samples, but with a gentler slope. Although it is possible to achieve an effective distance of around 7m, a higher 'v\_H\_S1' than 9.0m/s could better ensure an around 7m performance. According to the ICE curves and the sample points, at the lower velocity region (<9.0m/s), the effective distance is more easily influenced by other features. On the other hand, when the velocity is larger than 9.0m/s, it becomes more difficult to further improve the performance purely by a larger 'v\_H\_S1' or using different techniques (from the data perspective, modifying the values of the other features). Figure \ref{fig:pdp_1_women} shows the interactive contribution of 'a\_knee\_LD' with 'v\_H\_S1'. For a given value of 'v\_H\_S1', higher 'a\_knee\_LD' tend to lead to a longer effective distance, with an optimum between 150° and 160°. In addition, the right bottom part of the figure should be taken with caution, as it is purely extrapolation with no samples distributed in that region. 
Figure \ref{fig:ice_2_women} and Figure \ref{fig:pdp_2_women} study the influence of 'r\_stepDiff\_S21'. According to Figure \ref{fig:ice_2_women}, 'r\_stepDiff\_S21' is not a decisive factor to the effective distance, but a lower value of 'r\_stepDiff\_S21' seems better for achieving a longer effective distance. Figure \ref{fig:pdp_2_women} shows that given the velocity of the second last step ('v\_H\_S2'), a reduction of the step length is favored. The larger the feature 'v\_H\_S2' is, the more reduction of step length is favored. The non-convex contour in Figure \ref{fig:pdp_2_women} indicates that some interactions with other features exist.
Likewise, Figure \ref{fig:ice_3_women} exhibits a positive correlation between 'r\_stepDiff\_S32' and the effective distance. Moreover, according to Figure \ref{fig:pdp_3_women}, the larger the velocity of the third last step ('v\_H\_S3') is, the higher increase in step length should take for the following step (i.e. the larger value of 'r\_stepDiff\_S32'). There are also interactions with the other features in the non-convex region.

\section*{Discussion}

This study employs machine learning models to analyze the impact of long jump features on the effective distance of top-tier performances by both male and female athletes in three World Championship finals. The experiments were conducted separately on men's and women's samples, because of the gender difference in feature and performance distributions. The analysis relies mainly on two methods, i.e. SHAP analysis for identifying the most important features and ICE plots for evaluating the numeric impacts of the features. The key features for both genders have been identified, accompanied by an analysis of the interactions among these features.

For male athletes' jumps, The following features emerged as significant based on the experimental results:
\begin{itemize}
    \item The velocity-related features are identified as the most important features (see Figure \ref{fig:shap_global}), which are the same conclusions as indicated by the physics mechanism of long jump and many other literature~\cite{vcoh2017kinematic, coh1995kinematic, linthorne2008biomechanics, linthorne2005optimum}. The velocity influence is most prominent for the best men's jump (Figure \ref{fig:shap_best_men}), whose approach velocities of the last three steps overtake the other jumps by a large extend. On the other hand, the PDP and ICE plots show that given the close approach velocity values, the effective distances could still vary a lot, which indicates that other factors are also important contributors.
    \item The SHAP analysis also shows that the take-off angle significantly influences the take-off velocity, as reflected by the features ‘v\_H\_TO’, ‘v\_V\_TO’ and ‘v\_TO’. As noted by Tan~\cite{tan2000kinematics}, the conversion cost between vertical and horizontal kinetic energy results in an optimal take-off angle that is lower than the ideal 45°, which maximizes range in ideal projectile motion. Shang~\cite{shang2022research} further suggests that the optimal take-off angle lies between 35° and 40°, while Linthorne~\cite{linthorne2005optimum} provides a more comprehensive analysis, proposing an optimal range of 20° to 25°, which varies as a function of take-off velocity magnitude. 
    \item The analysis further identified the features related to the knee movement on board as the moderately important features for the men's samples (Figure \ref{fig:shap_global}), which is relevant to the process of transferring momentum from horizontal to vertical. A further analysis with the PDP and ICE plots in Figure \ref{fig:ice_pdp_men} reveals that the angle of 169° of 'a\_knee\_TD’ is a threshold for achieving better performance. This result can be interpreted using the spring-mass model of the take-off leg \cite{seyfarth1999dynamics, seyfarth2000optimum, muraki2005athletics}. Muraki et al. \cite{muraki2005athletics} suggested that a larger value of 'a\_knee\_TD’ is associated with greater spring strength. Supporting this, the calculations of Seyfarth et al. \cite{seyfarth2000optimum} indicate that the take-off distance decreases as 'a\_knee\_TD’ decreases from 170° to 160°, and further to 150°.
\end{itemize} 

The velocity-related features are also identified as the most important features for the female athletes' samples, but with less importance compared to the men athletes' samples (see Figure \ref{fig:shap_global}). This is mainly due to the fact that a large proportion of the women's samples achieved effective distances around 7m with different velocity values. Therefore, other features related to long jump techniques share more importance for women's samples. 
\begin{itemize}
    \item The first feature is the landing knee angle ('a\_knee\_LD'), which is suggested better to be between 150° and 160° (Figure \ref{fig:pdp_1_women}). Although landing-related features have been less extensively studied in the literature, their influence is evident due to their association with landing energy loss \cite{panoutsakopoulos20103d}. However, to the best of our knowledge, no quantitative analysis has been conducted to date.
    \item Two other features are the step length change percentage between the third last step and the second last step, and between the second last step and the last step ('r\_stepDiff\_S32' and 'r\_stepDiff\_S21'), which are related to the 'larger penultimate - shorter last step' technique \cite{panoutsakopoulos2021biomechanical, panoutsakopoulos2017gender}. These two features become important only when they have large magnitude values, which occurs in the three best women's jumps. Accordingly, they are the most important features for the best women's jump, which are related to the velocity improvement from the third last step to the second last step and velocity adjustment from the second last step to the last step in that jump. 
    \item Another particularly interesting finding is the gender difference in the identified influential features. Similar conclusions have been reported in the literature \cite{akl2014biomechanical, panoutsakopoulos2017gender, panoutsakopoulos2020biomechanical}, indicating that take-off technique tends to play a more critical role for male athletes, whereas approach technique appears to be more influential for female athletes.
\end{itemize}

Finally, the methods employed in this study are entirely data-driven, meaning the results and conclusions are inherently constrained by the dataset's distribution. This characteristic can be viewed as both an advantage and a limitation. If the goal is to derive generalizable insights about feature impacts, a large and diverse dataset that includes the full range of feature and target variable distributions would be necessary. Conversely, if the objective is to investigate feature effects within a specific, conditioned dataset (e.g., focusing on elite long jumpers rather than participants of all skill levels, or examining the performance of a single athlete), the data-driven methods utilized here are highly effective and straightforward to apply, enabling the discovery of nuanced feature influences. Therefore, despite the limited dataset size used in this study, the findings remain valuable, provided that appropriate caution is taken when extrapolating them to other samples or datasets (e.g., by assessing the distributions of the features and the target variable). Moreover, when applied to the training data of a single athlete, these methods can effectively identify and analyze the features most significantly impacting the athlete's peak performance, providing actionable insights to guide training strategies and enhance performance outcomes.

\section*{Conclusions and Future Work}
In conclusion, we combined quantile Random Forest and the XAI methods of SHAP, PDP and ICE to analyze the contribution of biomechanical features to the top performance of male and female athletes in the World Championships. Our method identified that besides velocity as the most contributing features for both men and women, the taking-off technique is more crucial for men while the approaching and landing technique is more important for women. Quantitative evaluation of feature importance, feature interactions as well as their quantitative influence to the effective distance have been presented. The findings are supported by a comparative discussion with other literature that studies with different methods.

There are mainly two limitations of this study. The first is that the dataset used in this work is medium size compared to other long jump studies, or small size compared to other machine learning studies. A larger dataset can help to ensure the conclusion more robust, for example, with a lower ratio of imputed feature values, and less interpolation in PDP/ICE plots. The second is that the the complicated feature interactions are not sufficiently addressed in interpreting the model, which can potentially influence the feature importance evaluation when strong interaction exists, e.g. between the horizontal velocities of the last few steps. Therefore, future studies could be conducted with a larger dataset, and by introducing other XAI methods that could better capture feature interactions, e.g. Accumulated
Local Effects (ALE) \cite{apley2020visualizing}, Visual INteraction Effects (VINE) \cite{britton2019vine} and Regional Effect Plots with implicit Interaction
Detection (REPID) \cite{herbinger2022repid}.

\section*{Data availability}
The raw data is sourced from reports\cite{Ref_report09, Ref_report17men, Ref_report17women, Ref_report18men, Ref_report18women} available on the official World Athletics website. We also provide our .csv version of the raw data as well as imputed data on the github project page \href{https://github.com/QGAN2019/Feature_interpretation_QRF}{https://github.com/QGAN2019/Feature\_interpretation\_QRF}. The code for practicing the experiments is also provided on the same project page.

\bibliography{sample}

\section*{Author contributions statement}

Q.G. prepared the data, developed the model, conducted the analysis, and wrote the manuscript. S.C. provided methodological guidance. M.A.E., S.M.N. and E.F. contributed to methodological guidance and supervised the experiments. O.J. provided input on early-stage experiments. All authors reviewed the manuscript. 

\section*{Additional information}

\subsection*{Competing interests}
The authors declare no competing interests.

\end{document}

\section*{Supplementary Information} % Optional: give the appendix a title

\vspace{1em}
{\LARGE \textbf{Feature Impact Analysis on Top Long-Jump Performances with Quantile Random Forest and Explainable AI Techniques} \par}
\vspace{1em}
\noindent\large Qi Gan$^{1,*}$, Stephan Clémençon$^{1}$, Mounîm A.El-Yacoubi$^{2}$, Sao Mai Nguyen$^{3}$, Eric Fenaux$^{4}$, Ons Jelassi$^{1}$

\vspace{1em}
\noindent\large $^{1}$LTCI, Télécom Paris, Institut Polytechnique de Paris, 91120 Palaiseau, France

\noindent\large $^{2}$SAMOVAR, Télécom SudParis, Institut Polytechnique de Paris, 91120 Palaiseau, France

\noindent\large $^{3}$U2IS, ENSTA Paris, Institut Polytechnique de Paris, 91120 Palaiseau, France

\noindent\large $^{4}$Ef-e-science, 75000 Paris, France

\noindent\large $^{*}$qi.gan@telecom-paris.fr

\appendix
\renewcommand{\thetable}{S\arabic{table}}  % Prefix tables with "S"
\renewcommand{\thefigure}{S\arabic{figure}}  % Prefix figures with "S"

\section{Feature explanation}
\noindent
\begin{table}[H]
\centering
\begin{tabular}{|p{3.5cm}|p{13cm}|}
\hline
\multicolumn{2}{|c|}{\textbf{Target features}}\\
\hline
d\_resEffe (m) & Effective distance (the real distance of jump). \\
\hline
d\_resOffi (m) & Official distance (measured from the take-off board).\\
\hline
\multicolumn{2}{|c|}{\textbf{Independent features}}\\
%\multicolumn{2}{|c|}{\textbf{Distance/length-related features}}\\
\hline
d\_step\_S3 (m) & The step length of the third last step.\\
\hline
d\_step\_S2 (m) & The step length of the second last step.\\
\hline
d\_step\_S1 (m) & The step length of the last step.\\
\hline
d\_loss\_TO (cm) & The front of take-off foot to the edge of take-off board, or 'd\_resEffe' minus 'd\_resOffi'. \\
\hline
d\_loss\_LD (m) & The distance between the first landing contact point and the final effective landing point. \\
\hline
d\_LD (m) & The horizontal distance between the center of mass (CoM) and the contact point at the first landing contact instant.\\
\hline
r\_stepDiff\_S32 (\%) & The step length change percentage from the third last step to the second last step.\\
\hline
r\_stepDiff\_S21 (\%) & The step length change percentage from the second last step to the last step.\\
\hline
%\multicolumn{2}{|c|}{\textbf{Time-related features}}\\
%\hline
t\_contact\_S3 (ms) & The contact time (foot on the ground) of the third last step.\\
\hline
t\_contact\_S2 (ms) & The contact time (foot on the ground) of the second last step.\\
\hline
t\_contact\_S1 (ms) & The contact time (foot on the ground) of the last step.\\
\hline
t\_flight\_S3 (ms) & The flight time (both feet in the air) of the third last step.\\
\hline
t\_flight\_S2 (ms) & The flight time (both feet in the air) of the second last step.\\
\hline
t\_flight\_S1 (ms) & The flight time (both feet in the air) of the last step.\\
\hline
t\_step\_S3 (ms) & The step time (contact time plus flight time) of the third last step. \\
\hline
t\_step\_S2 (ms) & The step time (contact time plus flight time) of the second last step.\\
\hline
t\_step\_S1 (ms) & The step time (contact time plus flight time) of the last step.\\
\hline
t\_TDO (s) & The duration between the foot touches the board and the foot leaves the board.\\
\hline
%\multicolumn{2}{|c|}{\textbf{Velocity-related features}}\\
%\hline
v\_H\_S3 (m/s) &  The mean horizontal velocity of CoM during the third last step.\\
\hline
v\_H\_S2 (m/s) & The mean horizontal velocity of CoM during the second last step.\\
\hline
v\_H\_S1 (m/s) & The mean horizontal velocity of CoM during the last step.\\
\hline
v\_H\_TO (m/s) & The horizontal velocity component of CoM at take-off.\\
\hline
v\_V\_TO (m/s) & The vertical velocity component of CoM at take-off.\\
\hline
a\_TO (°) & The angle of the velocity of CoM at take-off.\\
\hline
v\_TO (m/s) & The magnitude of the velocity of CoM at take-off.\\
\hline
v\_HDiff\_TDO (m/s) & The change of horizontal velocity from touchdown to take-off.\\
\hline
%\multicolumn{2}{|c|}{\textbf{Angle-related features}}\\
%\hline
a\_body\_TD (°) & The body inclination angle at touchdown.\\
\hline
a\_body\_TO (°) & The body inclination angle at take-off.\\
\hline
a\_trunk\_TD (°) & The trunk inclination angle at touchdown.\\
\hline
a\_trunk\_TO (°) & The trunk inclination angle at take-off.\\
\hline
a\_trunkTRot\_TDO (°) & The trunk rotation angle from touchdown to take-off.\\
\hline
a\_thigh\_TO (°) & The swinging thigh angle at the instant of take-off.\\
\hline
w\_thigh\_TDO (°/s) & The mean angular velocity of the swinging thigh from touchdown to take-off.\\
\hline
a\_knee\_TD (°) & The angle of the knee of the supporting leg at take-off.\\
\hline
a\_kneeMin\_TDO (°) & The minimum knee angle of the supporting leg from touchdown to take-off.\\
\hline
a\_kneeRange\_TDO (°) & The difference between 'a\_knee\_TD' and 'a\_kneeMin\_TDO'.\\
\hline
w\_knee\_TDO (°/s) & The mean angular velocity of the knee angle from touchdown to the minimum.\\
\hline
a\_hip\_LD (°) & The hip angle at the first landing instant.\\
\hline
a\_knee\_LD (°) & The knee angle at the first landing instant.\\
\hline
a\_trunk\_LD (°) & The trunk angle at the first landing instant.\\
\hline
%\multicolumn{2}{|c|}{\textbf{Other features}}\\
%\hline
Height (m) &  The height of the athlete found online.\\
\hline
Weight (kg) & The weight of the athlete found online.\\
\hline
h\_CMLower (cm) & The drop of the CoM from the start of the last step to the lowest during foot on the board.\\
\hline
\end{tabular}
\caption{\label{tab:feature_explanation}The explanations of the features.}
\end{table}

\section{Feature distributions}
\begin{figure}[H]
    \centering
    \includegraphics[width=0.9\linewidth]{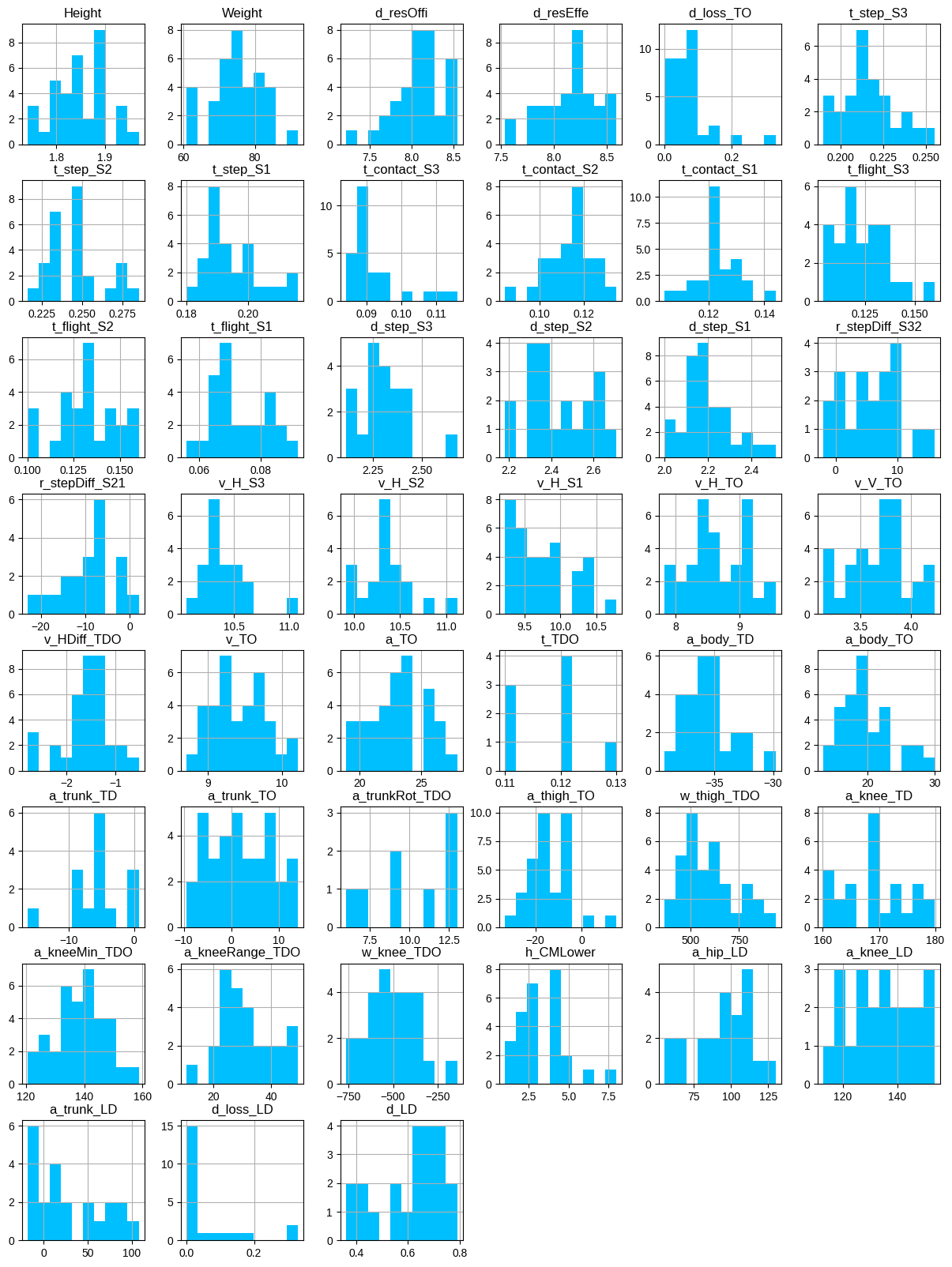}
    \caption{Histogram of features of \textit{male} athletes (before feature imputation).}
    \label{fig:feature_distribution_men}
\end{figure}

\begin{figure}[H]
    \centering
    \includegraphics[width=0.9\linewidth]{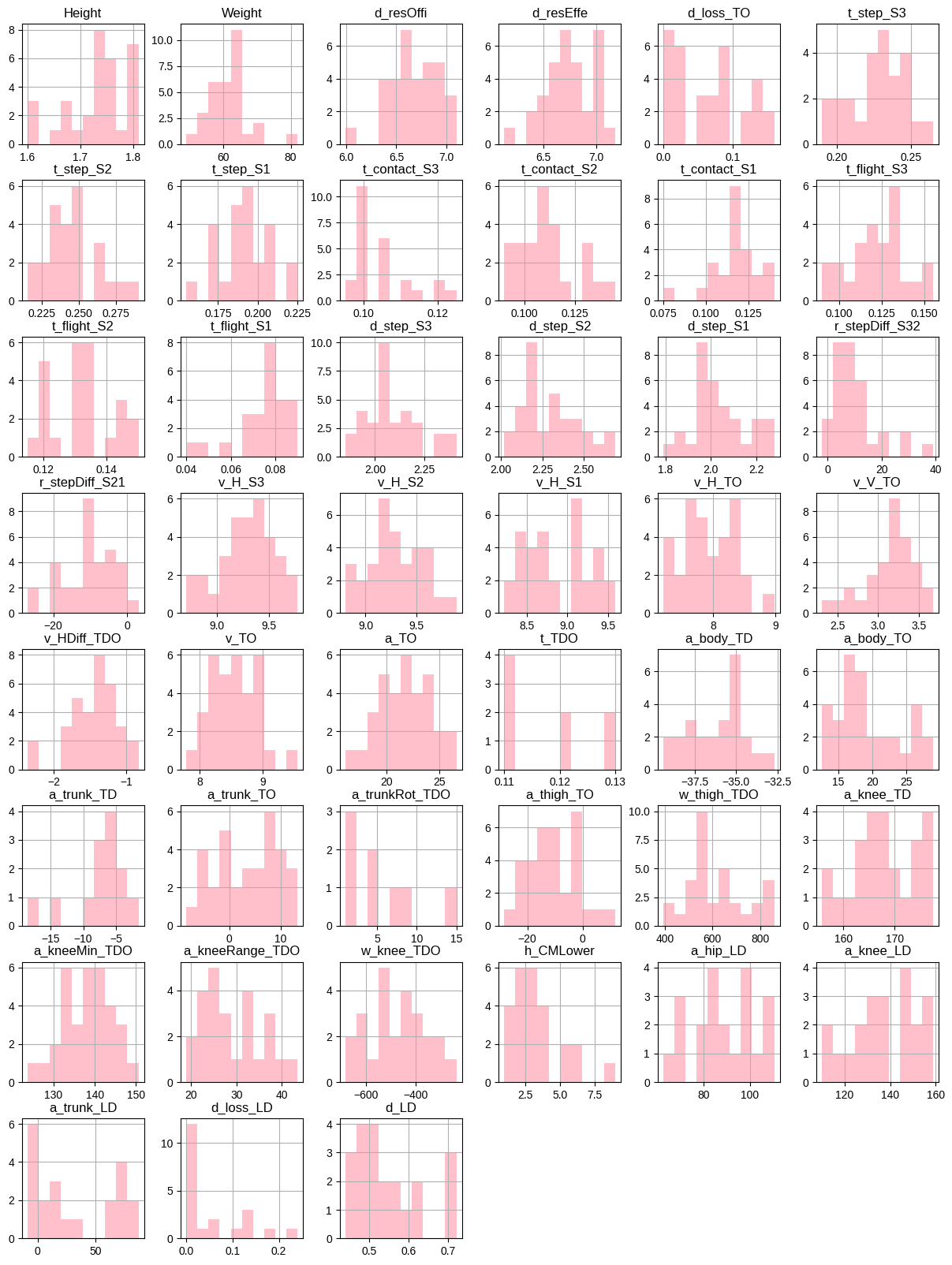}
    \caption{Histogram of features of \textit{female} athletes (before feature imputation).}
    \label{fig:feature_distribution_women}
\end{figure}

\newpage\section{Feature correlations}
\begin{figure}[H]
    \centering
    \includegraphics[width=1.0\linewidth]{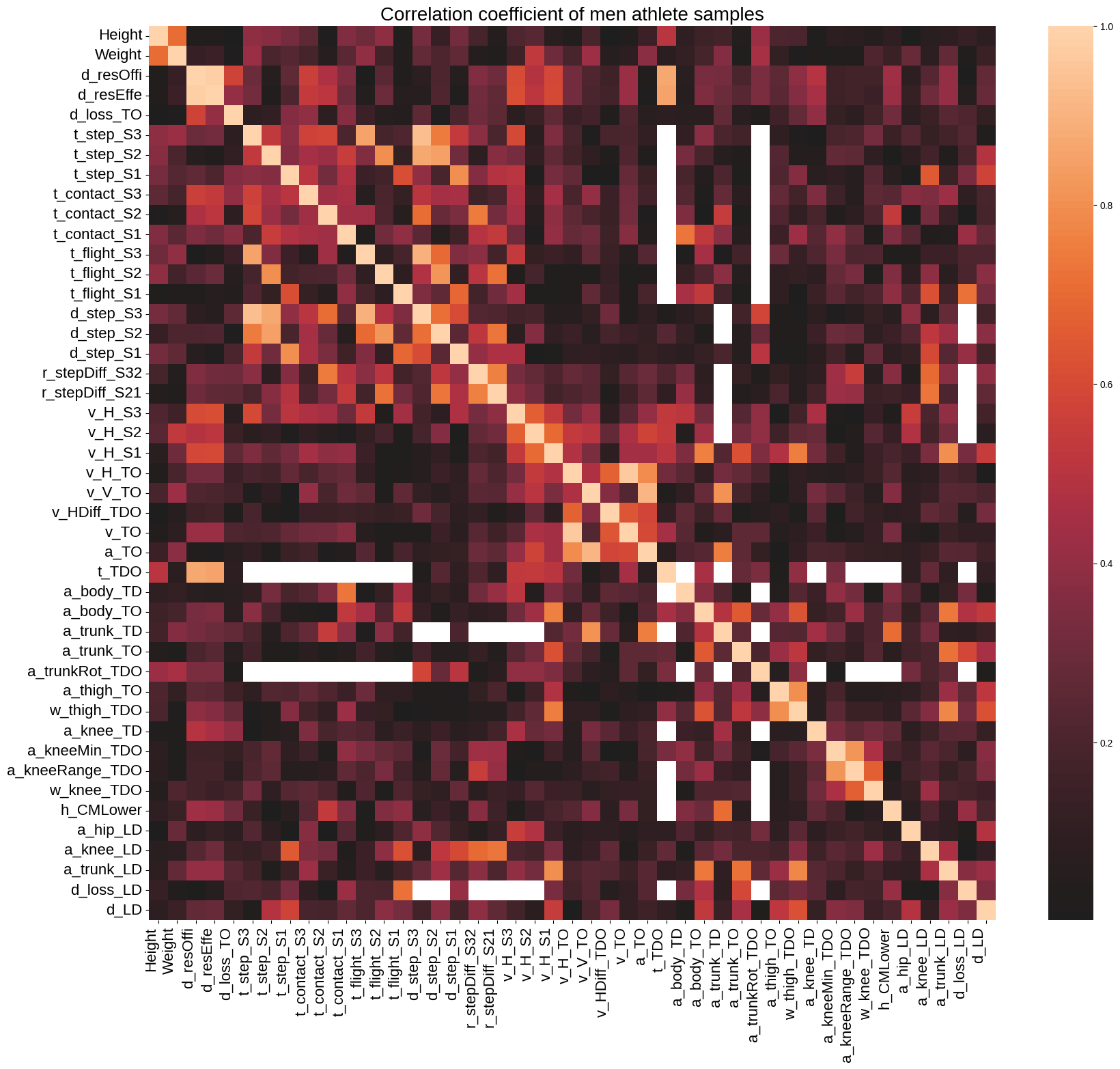}
    \caption{Correlation matrix of features for \textit{male} athletes (before feature imputation). Values represent the absolute correlation coefficients. White blanks indicate missing data, where no samples share common values for the corresponding feature pairs.}
    \label{fig:feature_corr_men}
\end{figure}

\begin{figure}[H]
    \centering
    \includegraphics[width=1.0\linewidth]{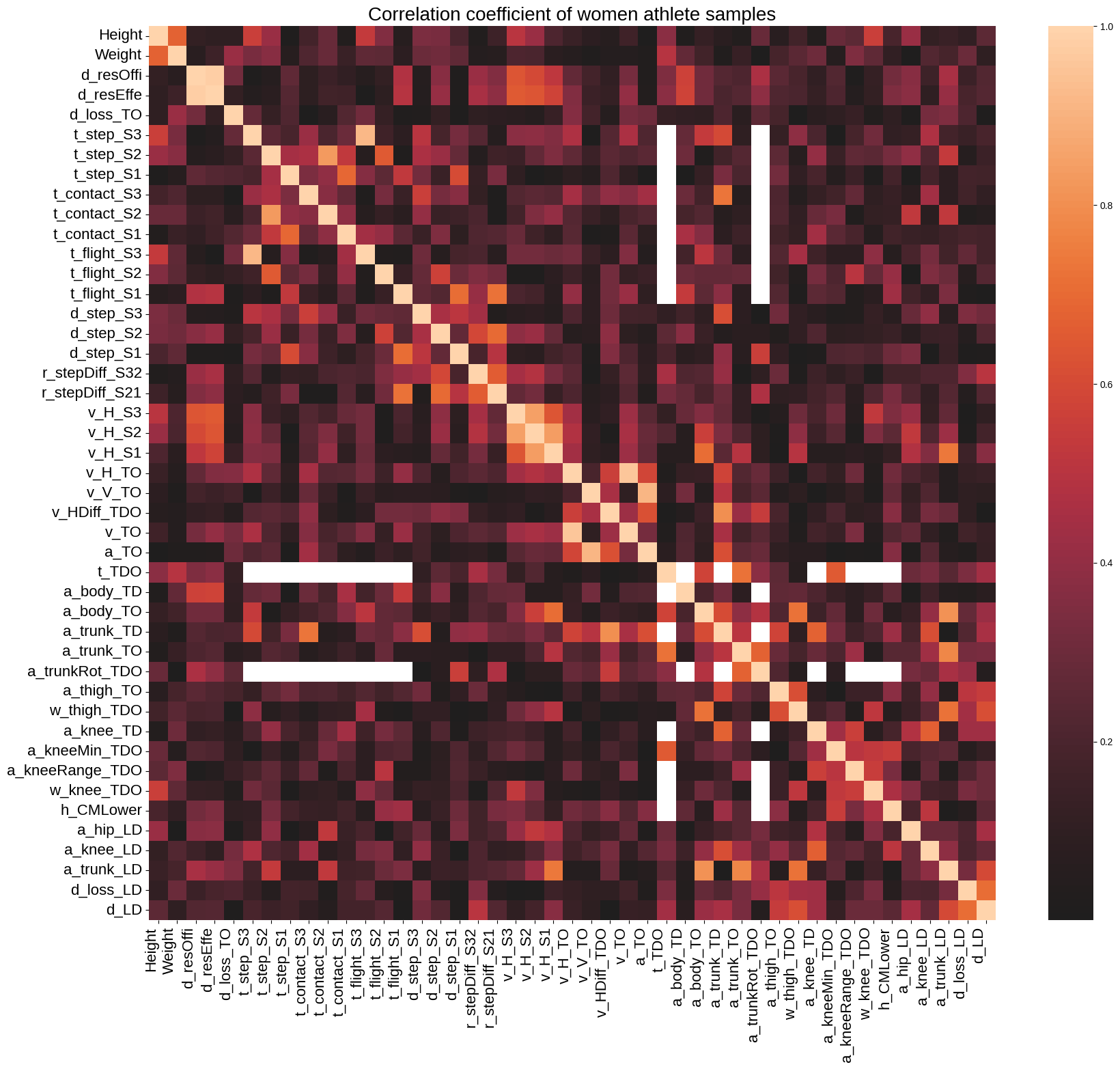}
    \caption{Correlation matrix of features for \textit{female} athletes (before feature imputation). Values represent the absolute correlation coefficients. White blanks indicate missing data, where no samples share common values for the corresponding feature pairs.}
    \label{fig:feature_corr_women}
\end{figure}

\newpage\section{Experimental results on the whole dataset}
In the main text, we trained and interpreted two separate regression models for male and female athletes. This approach not only aids in interpreting top performances for each gender but also better captures intra-gender variations. To support this choice, we compare the performance of a unified model trained on all athletes with that of the two gender-specific models.

To compare the predictive performances, we used standard Random Forest regressors\cite{breiman2001random} instead of their quantile variants. First, we applied the same feature imputation procedure as in the main text to the full dataset. Next, we performed feature selection using Lasso regression on the combined data as well as separately for male and female subsets. Finally, we trained a single model on the full dataset and compared its stratified performance with the two models trained separately by gender. The following sections present the experiments and results in detail.

\subsection{Feature imputation}
We evaluated the performance of four imputation methods using the same procedure described in the main text. The results are summarized in Table~\ref{tab:imputation_all}. Among the methods, the Random Forest-based iterative imputationcite\cite{breiman2001random, tang2017random} achieved the best evaluation performance and was therefore used in the subsequent experiments.

\begin{table}[ht]
    \centering
    \begin{tabular}{|l|c c c|}
        \hline
        Method & MSE$(\downarrow)$ & RMSE$(\downarrow)$ & $R^2$$(\uparrow)$ \\ 
        \hline
        Mean & 0.0592 & 0.243 & 0.895 \\
        \hline
        KNN & 0.0597 & 0.244 & 0.894 \\
        \hline
        Bayesian & 0.0566 & 0.238 & 0.899 \\
        \hline
        RF & \textbf{0.0526} & \textbf{0.229} & \textbf{0.906} \\
        \hline
    \end{tabular}
    \caption{\label{tab:imputation_all} Evaluation of four imputation methods using MSE, RMSE and $R^2$ on the whole dataset.}
\end{table}

\subsection{Feature selection}
Linear Lasso regression cross validation was used for feature selection, implemented with the \textit{scikit-learn} package\cite{pedregosa2011scikit} as in the main text. First, feature selection was performed on the full dataset using four-fold cross-validation to identify the optimal value of the regularization parameter \textit{alpha}. This process selected 10 features, with the best \textit{alpha} being 0.0307. The same procedure was then applied to the male athletes’ data, resulting in 13 selected features at \textit{alpha} = 0.0180. Lastly, for the female athletes’ data, 5 features were selected with an optimal \textit{alpha} of 0.0488. The selected features are summarized below:

\begin{itemize}
    \item Whole data (10 features): 'Height', 'Weight', 't\_contact\_S2', 't\_flight\_S2', 'v\_H\_S3', 'v\_H\_S2', 'v\_H\_S1', 'v\_V\_TO', 'v\_TO', 'Gender\_isFemale'.
    \item Male athletes' data (13 features): 'Height', 'd\_loss\_TO', 't\_contact\_S2', 't\_flight\_S2', 'd\_step\_S1', 'v\_H\_S3', 'v\_H\_S2', 'v\_H\_S1', 'v\_V\_TO', 'v\_TO', 'h\_CMLower', 'a\_knee\_LD', 'd\_loss\_LD'.
    \item Female athletes' data (5 features): 't\_flight\_S1', 'v\_H\_S3', 'v\_H\_S2', 'v\_H\_S1', 'a\_body\_TD'.
\end{itemize}

\subsection{Model evaluation}
We compared a Random Forest regression model trained on the entire dataset with models trained separately on male and female athlete data, using stratified 5-fold cross-validation based on gender. Within each stratified fold, an inner 4-fold cross-validation was conducted to select the optimal hyperparameters. In each outer fold, the model trained on the full training set was evaluated on the test set as a whole, as well as separately on the male and female subsets. In parallel, gender-specific models were trained using the corresponding gender's training data and feature sets, and evaluated on the respective gender-specific test sets. 

The results are presented in Table~\ref{tab:model_performance_comparison}. As shown, models trained on single-gender data outperform the model trained on the combined dataset, suggesting that gender-specific models better capture intra-gender variations. Additionally, the high $R^2$ value observed when training and evaluating on the combined dataset indicates that the model primarily captures inter-gender differences, rather than fine-grained variation within each gender.

\begin{table}[H]
    \centering
    \begin{tabular}{|l|l|c c c|}
        \hline
        Train data & Test data & MSE$(\downarrow)$ & RMSE$(\downarrow)$ & $R^2$$(\uparrow)$\\
        \hline
        \multirow{3}{*}{Combined} & Combined & 0.0450 & 0.212 & 0.920 \\
        \cline{2-5}
        & Men & 0.0404 & 0.199 & 0.396 \\
        \cline{2-5}
        & Women & 0.0496 & 0.220 & 0.036 \\
        \hline
        Men & Men & 0.0374 & 0.192 & 0.441 \\
        \hline
        Women & Women & 0.0254 & 0.158 & 0.503 \\
        \hline
    \end{tabular}
    \caption{\label{tab:model_performance_comparison} Model performance comparison with gender stratification. }
\end{table}

\subsection{Model interpretation}
We trained a Random Forest regression model on the combined dataset. Hyperparameter tuning was performed using 5-fold cross-validation, yielding the optimal values: 200 for 'n\_estimators', 6 for 'max\_features', and 9 for 'max\_depth'. The final model was then trained on the full dataset using these parameters.

\begin{figure}[H]
    %\centering  
    % First plot
    \begin{subfigure}{0.49\textwidth}
        \centering
        \includegraphics[width=\textwidth]{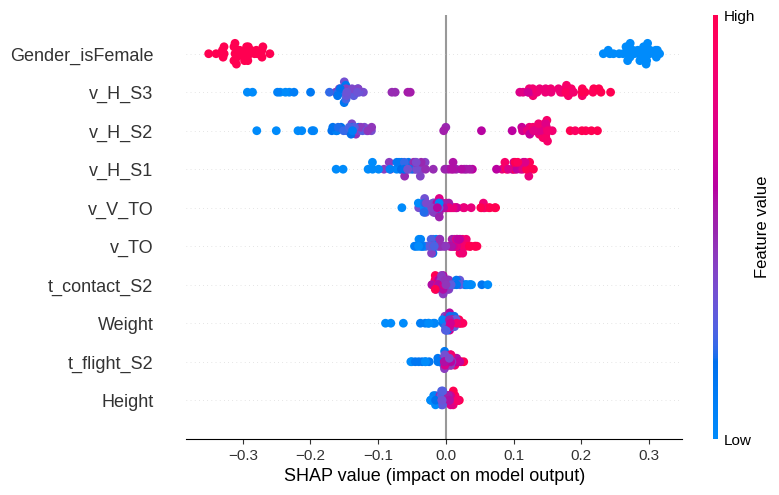}
        \caption{SHAP beeswarm plot on combined dataset (including both \textit{men} and \textit{women} athletes).}
        \label{fig:shap_beeswarm_all}
    \end{subfigure}
    %\vspace{1cm} % Space between the two plots
    % Second plot
    \begin{subfigure}{0.49\textwidth}
        \centering
        \includegraphics[width=\textwidth]{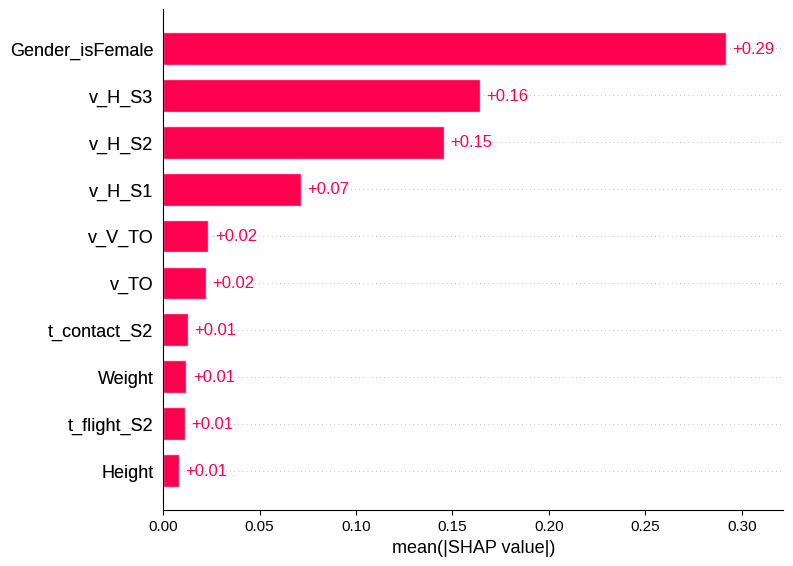}
        \caption{SHAP bar plot on combined dataset (including both \textit{men} and \textit{women} athletes).}
        \label{fig:shap_bar_all}
    \end{subfigure}
    \caption{SHAP analysis of the model trained on the combined data. }
    \label{fig:shap_all}
\end{figure}

Model interpretation was conducted using SHAP\cite{lundberg2017unified}, with the results shown in Figure~\ref{fig:shap_all}. The analysis reveals that gender is the most influential feature in the model's predictions. The next most important features are related to horizontal velocity, which also show remarkable differences between genders (see Figure~\ref{fig:feature_distribution_men} and Figure~\ref{fig:feature_distribution_women}). These findings suggest that the model primarily captures inter-gender variation, consistent with previous observations.

\bibliography{sample}